\documentclass[12pt, a4paper]{article}

\usepackage[english]{babel}

\usepackage[letterpaper,top=2cm,bottom=2cm,left=3cm,right=3cm,marginparwidth=1.75cm]{geometry}

\usepackage[dvipsnames]{xcolor}

\usepackage{amsmath}
\usepackage{amssymb}
\usepackage{graphicx}
\usepackage[colorlinks=true, allcolors=blue]{hyperref}
\usepackage{booktabs}
\usepackage{placeins}
\usepackage{subcaption}
\usepackage{ulem}

\usepackage{authblk}

\title{Explainable Global Wildfire Prediction Models using Graph Neural Networks}
\author[1]{Dayou Chen}
\author[2*]{Sibo Cheng}
\author[1]{Jinwei Hu}
\author[3]{Matthew Kasoar}
\author[1]{Rossella Arcucci}
\affil[1]{Department of Earth Science and Engineering, Imperial College London, London, United Kingdom}
\affil[2]{Data Science Institute, Imperial College London, London, United Kingdom}
\affil[3]{Department of Physics, Imperial College London, London, United Kingdom}
\affil[*]{Correspondence: sibo.cheng@imperial.ac.uk}

\begin{document}
\maketitle

\begin{abstract}
Wildfire prediction has become increasingly crucial due to the escalating impacts of climate change. Traditional CNN-based wildfire prediction models struggle with handling missing oceanic data and addressing the long-range dependencies across distant regions in meteorological data. In this paper, we introduce an innovative Graph Neural Network (GNN)-based model for global wildfire prediction. We propose a hybrid model that combines the spatial prowess of Graph Convolutional Networks (GCNs) with the temporal depth of Long Short-Term Memory (LSTM) networks. Our approach uniquely transforms global climate and wildfire data into a graph representation, addressing challenges such as null oceanic data locations and long-range dependencies inherent in traditional models. Benchmarking against established architectures using an unseen ensemble of JULES-INFERNO simulations, our model demonstrates superior predictive accuracy. Furthermore, we emphasise the model's explainability, unveiling potential wildfire correlation clusters through community detection and elucidating feature importance via Integrated Gradient analysis. Our findings not only advance the methodological domain of wildfire prediction but also underscore the importance of model transparency, offering valuable insights for stakeholders in wildfire management.
\end{abstract}


\noindent\begin{minipage}{\textwidth}
\section*{Main Notations}
\centering
\begin{tabular}{ll}
\toprule
\textbf{Notation} & \textbf{Description} \\
\midrule
\( G \) & Graph representation of the global surface data simulation\\
\( A \) & Adjacency matrix of the graph\\
\( \hat{A} \) & Normalised form of the adjacency matrix \\
\( \tilde{A} \) & Adjacency matrix with self-connection \\
\( a_{ij} \) & Strength of connection between nodes \( v_i \) and \( v_j \) \\
\( r \) & Wildfire Occurrance Correlation coefficient \\
\( E \) & Set of edges in the graph \\
\( e_{ij} \) & Edge between nodes \( v_i \) and \( v_j \) \\
\( N \) & Total number of nodes in the graph\\
\( V \) & Set of nodes in the graph \\
\( T_t \) & Temperature at time \( t \) \\
\( Hum_t \) & Humidity at time \( t \) \\
\( R_t \) & Rainfall at time \( t \) \\
\( L_t \) & Lightning at time \( t \) \\
\( P_t \) & Burnt area fraction at time \( t \) \\
\( P_{\zeta,t} \) & Fire snapshots for ensemble member \( \zeta \) at time \( t \) \\
\( X \) & Input Graph feature matrix \\
\( F \) & Mapping function of the proposed model \\
\( x \) & Feature vector for a node \\
\( Z \) & Hidden state feature matrix \\
\( I \) & Identity matrix \\
\( W \) & Learnable weight matrix of filter parameters \\
\( \sigma \) & Sigmoid activation function \\
\( f_t, i_t, o_t \) & Activations of the LSTM gates \\
\( \tilde{C}_t \) & Cell candidate value \\
\( C_t \) & Cell state at time \( t \) \\
\( b \) & Bias vectors \\
\( Q \) & Modularity of a partition \\
\( m \) & Number of links in the graph \\
\( k_i \) & Sum of the weights of the links attached to node \( i \) \\
\( c_i \) & Community to which node \( i \) is assigned \\
\( \delta(u, v) \) & Kronecker delta function: 1 if \( u = v \) and 0 otherwise \\
\( \mathcal{G} \) & Attribution of the \( i_{th} \) feature \\
\( \bar{X} \) & Baseline input for the Integrated Gradient method \\
\( \alpha \) & Variable scalar for Integrated Gradient, varies from 0 to 1 \\
\bottomrule
\end{tabular}
\end{minipage}

\section{Introduction}

Wildfires have been a subject of great interest due to their significant socioeconomic and environmental impacts~\cite{grillakis2022climate}. An average of 200,000 wildfires occur annually, affecting more than 1\% of the world's forests~\cite{pais2021deep}. Consequently, wildfire prediction has become a key interest. The frequency of global wildfires is intricately linked to factors such as land usage, vegetation type, and meteorological conditions. Risk factors that increase the likelihood of wildfires include arid land, hot and dry weather, and flammable vegetation types. As such, by leveraging data from verified natural environment models, it becomes possible to predict the risk of forest fires. To address this, several comprehensive simulation models, such as the Earth System Models~\cite{claussen2002earth} and Dynamic Global Vegetation Models (DGVM)~\cite{Prentice2013}, were developed for this purpose. An instance of a DGVM is the Joint UK Land Environment Simulator - INteractive Fire and Emissions algorithm for Natural envirOnments (JULES-INFERNO)~\cite{best2011joint, clark2011joint}. JULES-INFERNO simulates the area burned by fire and emissions over time, taking into account various geographical parameters including population density, land use, and weather conditions~\cite{burton2019representation, mangeon2016inferno}.

Despite the considerable advancements in wildfire prediction made possible by DGVM simulation models, the implementation of such DGVM simulations can be computationally expensive. DGVMs such as JULES-INFERNO have high computational costs due to the complexity of the differential equations used and high data dimensionality~\cite{lawrence2013storing}. For instance, predicting wildfire occurrences over a 100-year simulation scenario necessitates approximately 17 hours of computing time when using 32 threads on the JASMIN High-performance Computing (HPC) service~\cite{jasmin2023, lawrence2013storing}. Given these challenges, developing an efficient digital twin to function as a surrogate model and speed up the prediction process has become a research priority.

Many efforts have been devoted to the construction of surrogate models of DGVMs and other types of dynamical systems~\cite{zhu2022, cheng2022}. Among the explored methods, deep learning techniques, especially Long Short-Term Memory networks (LSTMs), have emerged as promising candidates. Research has demonstrated that LSTMs are particularly adept at handling dynamic simulations with long-term temporal correlations~\cite{graves2005framewise,cheng2023machine}. LSTM networks excel at learning long-term dependencies in input sequences due to their gate structures, which effectively prevent the vanishing gradient problem that occurs when processing long sequences~\cite{hochreiter1997}. In addition, Convolutional LSTM (Conv-LSTM), a variant of LSTM, is of key reference for our work. Compared to conventional LSTM, one major advantage of Conv-LSTM is its ability to capture spatial correlations. Conv-LSTM exhibited excellent performance in precipitation nowcasting, demonstrating its ability to predict geographical information~\cite{shi2015}.

In the task of wildfire prediction, Liang et al.~\cite{liang2019} developed an LSTM based model for predicting the scale of forest wildfires in Alberta, Canada. Radke et al.~\cite{Radke2019} devised FireCast, a dedicated 6-layer 2D Convolutional Neural Network (CNN), targeting areas with high near-future wildfire spread risk using historical fire records. Malik et al.~\cite{Malik2021} developed an ensemble model that integrates SVM and XGBoost for processing weather data, alongside CNN and LSTM techniques to handle vegetation and remote sensing image data snapshots, to forecast wildfire risks. Bhowmik~\cite{bhowmik2022multimodal} constructed a wildfire database from environmental and historical data, then formulated a specialised U-Convolutional-LSTM neural network, achieving a higher prediction accuracy than conventional CNNs. Cheng et al.~\cite{cheng2022} presented an innovative data-model integration approach for near real-time wildfire progression forecasting. Combining CNN-based Reduced-Order Modelling (ROM), LSTM, data assimilation, and error covariance tuning, they applied their methodology to large wildfire events in California from 2017 to 2020. Zhang et al.~\cite{zhang2022} introduced a deep learning-based surrogate model utilising Convolutional Auto Encoder (CAE) and Long Short-Term Memory (LSTM) model for sequence-to-sequence prediction of global wildfire status. Their approach compared Principal Component Analysis (PCA) \cite{wold1987principal,gong2020reactor} and CAE as ROM techniques. The model interprets global climate and wildfire data as image snapshots, and employs ROM methods to compress five selected geographical features used in JULES-INFERNO into their respective latent spaces. Subsequently, an LSTM network was employed to make multi-step predictions, achieving the predictive tasks with significantly reduced computational costs

However, the accuracy of previous surrogate models leaves room for improvement. Most CNN-based models follow a common approach in meteorological studies to treat the earth's surface data as image snapshots. This introduces a problem since it overlooks the fact that wildfires are exclusive to land and do not occur in oceans. As certain climate and wildfire data cannot be collected from areas like water surfaces, data at these locations assigned an arbitrary constant, for instance, zero~\cite{bhowmik2022multimodal, zhang2022}. Such a substitution practice introduces bias and potentially hinders the predictive capacity of neural networks~\cite{kang2013prevention}.

Furthermore, CNN based architectures must be tailored to specific tasks, specifically, the size of the convolutional kernels at each layer must be determined before training to consider the resolution, width and depth of the input data. As a consequence, CNN-based networks with limited layers may lack the capacity to capture long-range dependencies due to ineffective architectural designs for specific tasks~\cite{romero2022flexconv, knigge2023modelling}. At the heart of CAE is the convolution operation, which processes input data using local receptive fields, during which the new features of a pixel are computed from its neighbouring pixels, dictated by the size of the convolutional kernel~\cite{Rawat2017}. The pixels correspond to proximate locations on the earth's surface. Hence, locations significantly apart from each other may not fall within the same receptive fields, making it difficult for the convolutional layers to capture their correlation directly. In real-world meteorological data, due to phenomena like atmospheric Rossby waves and ocean currents, locations separated by vast distances, even oceans, can exhibit strong climatic correlations~\cite{MOCK2014ElNino, GENT2013609, lau1997}. Therefore, a more advanced technique is required to capture the long-range feature interactions in global surface wildfire data.

In recent years, the application of Graph Neural Network (GNN) has shown promise in the field of geographical information forecasting. Keisler~\cite{keisler2022forecasting} successfully used GNN to forecast complex 3D atmospheric information, demonstrating the efficacy of GNN for modelling the dynamics of physical systems in a data-driven manner. Cachay et al.~\cite{cachay2021graph} applied a spatiotemporal GNN model to predict El Niño-Southern Oscillation (ENSO) events, achieving finer granularity and better predictive skill than previous state-of-the-art methods. A graph modelling approach also offers a potential solution to address the issue of null data locations. Moreover, explicit modelling of information flow through graph edges provided more interpretable prediction results, which is crucial in wildfire prediction models as it offers transparency, builds trust for stakeholders, and facilitates better-informed decision-making in critical situations.

Despite promising applications of GNN in forecasting geographical information, to our knowledge, a GNN-based model for global wildfire prediction is yet to be developed. This paper proposes a novel explainable approach for wildfire prediction that leverages GNN in conjunction with LSTM, utilising a comprehensive dataset with climate and wildfire data extracted from JULES-INFERNO simulations. Our work proposes a novel method of representing the global climate and wildfire data as a graph, which is then processed using a Graph Convolutional Network (GCN)~\cite{kipf2017semisupervised}and an LSTM network for improved prediction accuracy and computational efficiency. The graph formulation is a correlation-based weighted graph, addressing the limitations regarding null oceanic data and long-range dependencies. This model not only boasts better predictive capabilities than existing models, but also provides a higher degree of interpretability. The efficacy of our model is validated on an unseen ensemble of 30-year JULES-INFERNO simulations to predict the monthly fire burnt area, benchmarking against LSTM, CAE-LSTM, and Conv-LSTM using various metrics including Mean Square Error (MSE).
To summarise, the main contributions of this study are as follows.
\begin{itemize}
    \item \textbf{Hybrid Model Innovation}: Our method integrates GCN with LSTM, introducing a GNN-based model for global wildfire prediction, presenting a methodological advancement in wildfire prediction.
    \item \textbf{Data Transformation}: We devise a novel graph representation of global climate and wildfire data, effectively addressing challenges including null oceanic data and long-range dependencies.
    \item \textbf{Enhanced Performance}: Our model offers superior predictive accuracy and increased interpretability compared to existing approaches. Using an unseen ensemble of 30-year JULES-INFERNO simulations, we rigorously benchmark our model against established architectures, validating its efficacy.
\end{itemize}

The structure of this paper is as follows: Section~\ref{sec:dataset} details the JULES-INFERNO simulations used as our dataset. Section~\ref{sec:relatedwork} we delve into deep learning-based techniques with a focus on CAE-LSTM and ConvLSTM. Our proposed methodology, combining Graph Convolutional Layer (GCL) and LSTM network, is explained in Section~\ref{sec:methodology}. The numerical results, assessed using metrics such as MSE, RRMSE, SSIM, and PSNR, are presented in Section~\ref{sec:results}. Section~\ref{sec:explainability} delves into the interpretability aspects of our model, covering community detection, and feature and node attributions. The paper concludes with a summary in Section~\ref{sec:conclusion}.

\section{Dataset}\label{sec:dataset}

\begin{figure}[htb]
\centering
\includegraphics[width=1.0\linewidth]{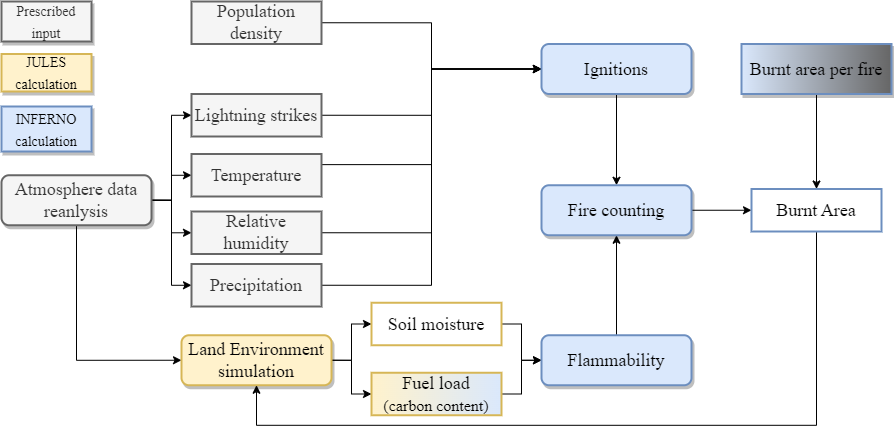}
\caption{\label{fig:JULES}The JULES-INFERNO process overview}
\end{figure}

The Joint UK Land Environment Simulator (JULES) models land vegetation, carbon levels, and water cycles using computational physical models. The INteractive Fire and Emissions algorithm for Natural envirOnments (INFERNO) calculates fire risk based on variables such as soil moisture, fuel density, and weather conditions, while also factoring in ignition sources such as lightning strikes~\cite{clark2011joint}. JULES, when paired with the INFERNO fire scheme, can predict areas burnt by fires, carbon emissions, and vegetation mortality~\cite{mangeon2016inferno}. Additionally, the system can estimate the emission of atmospheric aerosols and trace gases based on vegetation factors~\cite{  burton2019representation, mangeon2016inferno}. Figure~\ref{fig:JULES} depicts the structure and mechanism of the parameterisation process of JULES-INFERNO, highlighting the key input variables and components.
This study focuses on four meteorological boundary conditions: temperature \(T\), humidity \(Hum\), rainfall \(R\), and lightning \(L\). Additionally, it examines the predicted burnt area fraction \(P\) derived from the JULES-INFERNO model as input data. Five ensemble members of JULES-INFERNO simulations were utilised, driven by the meteorological conditions of the last glacial maximum (LGM) scenario from the FireMIP project~\cite{rabin2017fire}. This scenario uses a detrended re-analysis of atmospheric boundary conditions from 1961-1990, adjusted to align with the global average climate during the LGM. Monthly averages for each meteorological boundary condition were gathered, spanning from 1961-1990 for each simulation. Consequently, 360 data points for each parameter were available, referred to as temperature \(T_t\), \(t = 1, 2, ..., 360\), humidity \(Hum_t\), rainfall \(R_t\), and lightning \(L_t\). JULES-INFERNO was implemented at a resolution of 1.25° latitude by 1.875° longitude, with each snapshot spanning 144 latitude units and 192 longitude units, resulting in a snapshot size of \( 144 \times 192 \).

The predicted wildfire burned areas by each of the five ensemble members of JULES-INFERNO are designated as \(P_{\zeta}\), \(\zeta = 1, 2, ..., 5\). All ensemble members were simulated for the same nominal timeframe (1961 to 1990) and with the same detrended meteorological conditions. Still, different initial internal states were implemented to cover a variety of model internal variabilities. As illustrated in Table~\ref{tab:ensemble_members}, the initial internal state for \(P_1\) was randomised, with subsequent experiments' initial internal states being the final internal states of the preceding one. The model's output for the burned area was only diagnosed over latitudes featuring non-zero land cover. As a result, the number of latitude units covered by the fire snapshots \(P_{\zeta,t}\), \(\zeta = 1, 2, ..., 5\), \(t = 1, 2, ..., 360\) was reduced to 112, yielding a snapshot size of \( 112 \times 192 \). As our work focused on the wildfire predictions, we shifted and clipped the meteorological data to align with the wildfire burnt area data shape.

\begin{table}[htb]
\centering
\begin{tabular}{lcc}
\toprule
\textbf{Ensemble Member} & \textbf{Initial Internal State} & \textbf{Source of Initial State} \\
\midrule
\(P_1\) & Randomised & - \\
\(P_2\) & Based on \(P_1\) & Final state of \(P_1\) \\
\(P_3\) & Based on \(P_2\) & Final state of \(P_2\) \\
\(P_4\) & Based on \(P_3\) & Final state of \(P_3\) \\
\(P_5\) & Based on \(P_4\) & Final state of \(P_4\) \\
\bottomrule
\end{tabular}
\caption{Initial internal states for each ensemble member of the JULES-INFERNO simulation.}
\label{tab:ensemble_members}
\end{table}

In later experiments, \(P_1\), \(P_2\), \(P_3\), and a portion of \(P_5\) formed the training set, with another part of \(P_5\) serving as the validation set. \(P_4\) was used as the test set for all models constructed in this study. Furthermore, to avoid the negative impacts of outliers in the training data, all training and test sets (climate and fire variables) were standardised by applying Equation~\ref{eq:normalisation}, which normalises the data to the range [0, 1].

\begin{equation}
P_{\zeta}' = \frac{P_{\zeta} - \min(P_{\zeta})}{\max(P_{\zeta}) - \min(P_{\zeta})}, \zeta = 1,2,...,5
\label{eq:normalisation}
\end{equation}

\section{Related Work}\label{sec:relatedwork}

Deep learning-based approaches, primarily CNN- and RNN-based networks and their variations, are prominently used in the prediction and understanding of various temporal sequences. This section delves deeper into the mathematical and theoretical foundations of CAE-LSTM and ConvLSTM.

\subsection{Convolutional Auto Encoder with LSTM (CAE-LSTM)}

CAE-LSTM~\cite{maulik2021reduced,cheng2022data,xiao2019domain} models merge the strengths of Convolutional Auto Encoders (CAE) for spatial feature extraction and dimension reduction with LSTMs for temporal sequence prediction. Following the recent work of~\cite{zhang2022}, we consider the raw data modelled as an \( n \)-dimensional vector, \( U = \{u_i\}_{i=1}^{n} \). Post dimensionality reduction, the compressed variable emerges as \( \Tilde{U} = \{ \tilde{u_i}\}_{i=1}^{k} \), with \( k < n \).

\begin{figure}[ht]
    \centering
    \includegraphics[width=0.5\linewidth]{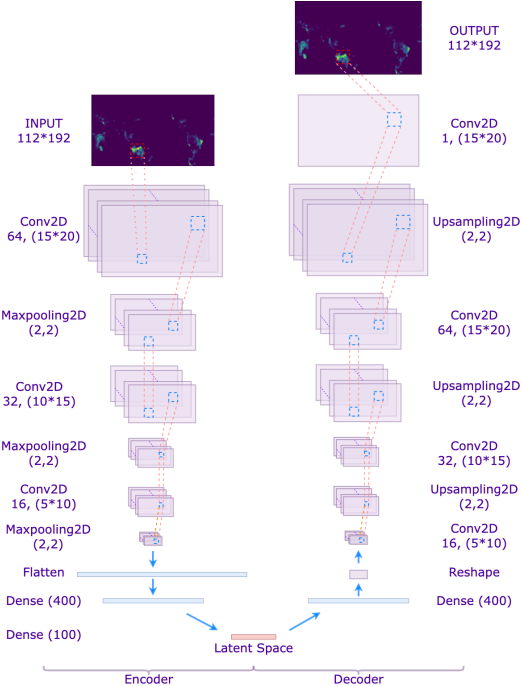}
    \caption{CAE structure with a latent space dimension of 100~\cite{zhang2022}}
    \label{fig:CAE_LSTM_structure}
\end{figure}

The encoder, depicted in Figure~\ref{fig:CAE_LSTM_structure}, incorporates three convolutional layers, three max pooling layers, and two fully connected layers. The decoder, in contrast, is assembled with four convolutional layers, three up-sampling layers, and one fully connected layer. Each meteorological snapshot, characterised by latitude and longitude, is considered an image for processing purposes. The mathematical operations for the implemented CAE encoder can be described as:

\begin{equation}
\begin{cases}
O_l = f_l (W_l \otimes U + b_l), & \text{for } l = 1 \\
O_l = f_l (W_l \otimes O'_{l-1} + b_l), & \text{for } l = 2,3
\end{cases} 
\end{equation}

\begin{equation}
O'_{l-1} = \text{Maxpooling}(O_l), \text{for } l = 1,2,3,
\end{equation}

where \( O_l \) is the output of the \( l_{th} \) convolutional layer, \( W_l \) and \( b_l \) represent the weights and biases for layer \( l \) respectively, and \( f_l (\cdot) \) signifies the activation function. The operation of the decoder is similar to the one of the encoder with deconvolutional layers.

Following the encoding process, the latent states, which are the compressed versions of the input climate and fire data, are passed to the LSTM for sequence-to-sequence prediction. The LSTM predicts future latent states which are then decoded using the trained CAE's decoder, converting them back to the original data dimensions\cite{zhang2022}. Dimension reduction and feature extraction approaches using CNNs have been widely adopt in wildfire modelling~\cite{zhang2022, cheng2022parameter, ban2020near}

\subsection{Convolutional LSTM (ConvLSTM)}

The ConvLSTM~\cite{shi2015}, an LSTM variant, uniquely integrates convolution operations into the LSTM structure, enabling it to efficiently manage spatial-temporal data. While LSTMs excel at managing time series data, the convolution operation within ConvLSTM aids in capturing spatial correlations, rendering it ideal for tasks like precipitation nowcasting and similar meteorological data forecasting applications. 

In ConvLSTM, the input, forget, and output gates, alongside the cell states, are all 3-dimensional tensors. The state update processes integrate convolutions, preserving the spatial structure, which can be defined as:
\begin{align}
i_t &= \sigma(W_{xi} \otimes X_t + W_{hi} \otimes Z_{t-1} + b_i) \\
f_t &= \sigma(W_{xf} \otimes X_t + W_{hf} \otimes Z_{t-1} + b_f) \\
o_t &= \sigma(W_{xo} \otimes X_t + W_{ho} \otimes Z_{t-1} + b_o) \\
\tilde{C}_t &= \tanh(W_{xc} \otimes X_t + W_{hc} \otimes Z_{t-1} + b_c) \\
C_t &= f_t \odot C_{t-1} + i_t \odot \tilde{C}_t \\
Z_t &= o_t \odot \tanh(C_t)
\end{align}

where \( \otimes \) denotes the convolution operation, \( \sigma \) is the sigmoid activation function, and \( \tanh \) represents the hyperbolic tangent function. The variables \( i_t \), \( f_t \), and \( o_t \) are the input, forget, and output gates, respectively, which control the flow of information in and out of the memory cell. The term \( \tilde{C}_t \) represents the candidate cell state, \( C_t \) is the cell state, and \( Z_t \) denotes the hidden state or the output of the ConvLSTM cell. The convolutional kernels \( W_{xi} \), \( W_{xf} \), \( W_{xo} \), and \( W_{xc} \) are applied to the input feature map \( X_t \), and the kernels \( W_{hi} \), \( W_{hf} \), \( W_{ho} \), and \( W_{hc} \) are applied to the previous hidden state \( Z_{t-1} \). Additionally, \( b_i \), \( b_f \), \( b_o \), and \( b_c \) represent the bias terms for the input, forget, output gates, and cell state candidate respectively.

This architecture allows ConvLSTM to process meteorological data with inherent spatial structures. Empirical tests demonstrated that ConvLSTM excels in capturing spatiotemporal correlations, consistently outpacing pure LSTM and conventional nowcasting methods~\cite{shi2015}. Mu et. al effectively utilised ConvLSTM for ENSO prediction by formulating ENSO as a spatialtemporal problem~\cite{EnsoConvLSTM2019}.

It is crucial to reiterate the underlying limitations of entirely CNN-based models in this domain. Specifically, the general practice of representing the earth's surface wildifre data as image snapshots may introduce biases as a large portion of the surface data contain missing values that must be filled with an arbitrary constant, such as zero. Additionally, the inherent architectural rigidity in designing CNNs, can hinder the capture of long-range dependencies, especially considering the vast and varied spatial correlations present in meteorological data~\cite{romero2022flexconv, knigge2023modelling}. Such complexities emphasise the need for more flexible and comprehensive approaches to understanding global wildfire dynamics.

\section{Graph neural network for wildfire prediction: Methodology}\label{sec:methodology}
To address the challenges of long-range dependencies and absent data, we propose a Graph Neural Network (GNN) model to surrogate and analyse global surface data simulation pixels from the JULES-INFERNO results. Each simulation pixel is conceptualised as a node in the graph, with the edges determined based on correlation coefficients between nodes. The methodology section commences with a detailed exposition on graph formulation, showing the mathematical representation of a graph and its components. Subsequently, we present the model structure, which integrates the Graph Convolutional Layer (GCL) and the LSTM network, each tailored to capture spatial and temporal dependencies, respectively. The intricacies of the GCL and the LSTM network are explored, followed by a comprehensive description of both layer's role in processing sequences of hidden states. Lastly, we present the experimental framework, evaluation metrics and three baseline models for comparative analysis.
\subsection{Graph Formulation}
Graph theory forms the core of GNNs. A graph or network is a means of representing data through a set of nodes a set of edges that define the pairwise relationships among the corresponding nodes. Every node connected to another node by an edge is considered its neighbour. These connections materialise in the model as an adjacency matrix. Formally, a graph \( G \) can be defined as \( G = (V, E) \) where \( V = \{V_i\}^N_{i=1} \) represents the set of nodes and \( E = \{e_{ij}\} \) represents the set of edges. Here, \( N \) is the total number of nodes and \( e_{ij} \) the pairwise connection between \( v_i \) and \( v_j \). \( N(v) = \{ u \in V | (v, u) \in E \} \) denotes the neighbourhood of nodes connecting to node \( v \). The connections of all nodes are described by the adjacency matrix \( A \in \mathbb{R}^{N\times N} \), its element \( a_{ij} \) is a float number specifying the proximity between nodes \( v_i \) and \( v_j \). Each node \( v \) contains a feature vector \( x \in \mathbb{R}^D \), where \( D \) is number of features. The graph feature matrix formed by all feature vectors is denoted as \( X \in \mathbb{R}^{N\times D} \). 

In response to the limitation of absent oceanic data, we introduce a land mask to the global JULES-INFERNO simulation data, restricting all data to solely encompass the earth's land surface. As a result, the dimensions of both climate and each wildfire ensemble member transit from images of \( 112 \times 192 \) to column matrices with length 7771, echoing an \( N \) of 7771 terrestrial data collection nodes. These nodes are comparable to localised weather and wildfire data collection stations. Following the work of~\cite{zhang2022}, the Temperature \( T_t \), humidity \( Hum_t \), rainfall \( R_t \), and lightning \( L_t \), coupled with the burnt area fraction \( P_t \), are gathered at each node. For any given time \( t \) and node \( i \), the feature vector is expressed as:
\begin{equation}
X^{(i)}_t = [T_t, Hum_t, R_t, L_t, P_t] 
\label{eq:feature_matrix}
\end{equation}
In the context of this study, we formulate the graph by treating the global surface data simulation pixels of the JULES-INFERNO simulation results as nodes and generate edges based on the correlation coefficients (\(r\)) between two nodes. Utilising correlation coefficients to represent edges enables the capture of long-range dependencies across distant regions. These correlations are calculated from the wildfire burned areas across a timeseries for all training ensemble members, namely, \(P_1\), \(P_2\), \(P_3\), and \(P_5\), each spanning 30 years. We set the cut-off thresholds at the 10\% quantiles of the \(r\) calculated from all nodes, further reducing the total number of edges. We ensured that this treatment provides maximum data compression while maintaining sufficient graph information and structure. The resulting \(r\) get allocated to the elements of the adjacency matrix, \( a_{ij} \), a float ranging between 0 and 1. This value quantifies the connection strength between two nodes, \( v_i \) and \( v_j \), with 1 indicating a robust link and 0 implying the absence of a connection. The process can be mathematically described as:
\begin{equation}
a_{ij} = 
\begin{cases} 
r(v_i, v_j) & \text{if } r(v_i, v_j) > 0.51 \text{ and } i \neq j, \\
0 & \text{otherwise}, 
\end{cases}
\label{eq:adjancency}
\end{equation}
where 0.51 is the 10\% quantile of all \(r\) calculated.

\subsection{Model structure}

\begin{figure}[ht]
\centering
\includegraphics[width=0.8\linewidth]{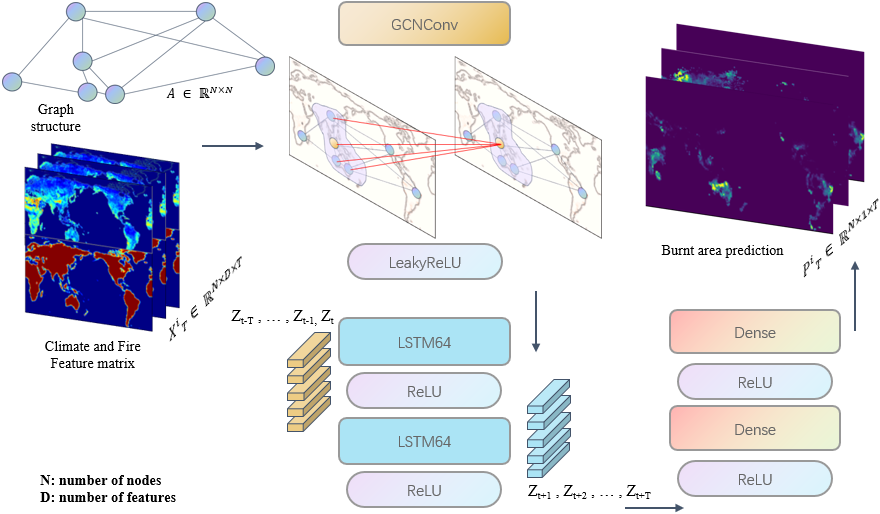}
\caption{\label{fig:model_structure}The GCN-LSTM model structure}
\end{figure}

In our research, we introduce a GNN model that prominently incorporates the Graph Convolutional Layer (GCL) and the LSTM network as the primary components of the architectural framework. The schematic representation of this network structure is illustrated in Figure~\ref{fig:model_structure}. While the GCL captures spatial dependencies within the nodes, the LSTM network derives the temporal dependencies. Collectively, this GNN model is engineered to execute a multi-step spatiotemporal prediction task. For this context, the input feature matrix is characterised by \( D \) features spanning \( T \) timesteps, encapsulated as a three-dimensional tensor \( \mathbf{X}_T \in \mathbb{R}^{N \times D \times T} \), where \( \mathbf{X}_T = [X_{t-T}, ... , X_{t-1}, X_t] \). The prediction process can be conceptualised as learning a mapping function \( F \) on the global wildfire correlation graph structure \( G \) and the feature matrix \( X \). Therefore, the prediction task can be formalised as:
\begin{equation}
[P_{t+1}, P_{t+2}, \ldots, P_{t+T} ] = F \left( G; (X_{t-T}, \ldots, X_{t-1}, X_t) \right)
\end{equation}
where \( T \) denotes the duration, in months, for which predictions are sought.

\subsection{Graph Convolutional Layer}

 Models based on graph convolution can be broadly divided into the spatial and the spectral approach. Spectral models operate in the graph's frequency domain, requiring eigendecomposition of the graph Laplacian, a computationally expensive operation ~\cite{Heidari2022_GCN}. In contrast, spatial methods operate directly on nodes and their immediate neighbourhoods, aggregating information from adjacent nodes. A specific type of spatial model, the Graph Convolutional Networks (GCNs) introduced by Kipf and Welling in 2016~\cite{kipf2017semisupervised}, simplifies the graph convolution using the first-order approximation of Chebyshev polynomials, avoiding the need for eigendecomposition of the Laplacian matrix, leading to a more computationally efficient process. The aggregation process of the GCN can be formalised as:
 
\begin{equation}
H^{(k)} = \sigma(\hat{A}H^{(k-1)}W^{(k)}) ,
\end{equation}
where \( \hat{A} \) is the normalised form of the adjacency matrix obtained by:
\begin{equation}
\hat{A} = \tilde{D}^{-1/2}\tilde{A}\tilde{D}^{-1/2} ,
\end{equation}
\begin{equation}
\tilde{A} = A + I ,
\end{equation}

where \( \tilde{A} \) is the adjacency matrix with self-connection, \( I \in \mathbb{R}^{N\times N} \) signifies the identity matrix, \( k\) represents the layer index in the graph convolutional network, with \( H^{(k)} \in \mathbb{R}^{N\times D_k} \)  denoting the feature matrix at the \( k_{th}\) layer,  \( W^{(k)} \in \mathbb{R}^{D_{k-1}\times D_k} \) represents a learnable matrix of filter parameters at the \( k_{th}\) layer. As we employ a single-layer GCN to perform spatial convolution, this process can be further simplified as:
\begin{equation}
Z = \sigma(\hat{A}XW) ,
\end{equation}
where \( X \in \mathbb{R}^{N\times D} \) is the input feature matrix, as defined in Equation~\ref{eq:feature_matrix}. In recent studies, spatial models have received more attention owing to their computational efficiency against their spectral counterparts~\cite{Heidari2022_GCN}. It is on these merits that the spatial-based GCN has been selected as the principal component of our graph convolution layer.

\subsection{Long Short-Term Memory (LSTM) Layer}

Following the extraction of spatial relationships using the Graph Convolutional Layer, the LSTM layer is employed to capture the temporal dependencies in the data. Given the hidden states \( Z \) derived from the GCN, the LSTM processes these states in a sequential manner. The outputs from the Graph Convolutional Layer can be represented as a sequence of hidden states \( Z_{t-T}, \ldots, Z_{t-1}, Z_t \). These states serve as the input to the LSTM layer, which is tasked with predicting the subsequent sequence \( Z_{t+1}, Z_{t+2}, \ldots, Z_{t+T} \). The core computations governing the LSTM updates are:
\begin{align}
f_t &= \sigma(W_f \cdot [Z_t, Z_{t-1}] + b_f), \\
i_t &= \sigma(W_i \cdot [Z_t, Z_{t-1}] + b_i), \\
\tilde{C}_t &= \tanh(W_C \cdot [Z_t, Z_{t-1}] + b_C), \\
C_t &= f_t \times C_{t-1} + i_t \times \tilde{C}_t, \\
o_t &= \sigma(W_o \cdot [Z_t, Z_{t-1}] + b_o), \\
Z_t &= o_t \times \tanh(C_t),
\end{align}

where, \( Z_t \) is the hidden state at time \( t \), the matrices \( W_f, W_i, W_C, \) and \( W_o \) are the learnable parameters of the LSTM unit, they transform the input data and hidden states for the LSTM’s gates and cell state updates, the definition of the rest of the parameters can be found in Section~\ref{sec:relatedwork}.

\subsection{Experiment Setup}

In our experimental framework, we benchmark the performance of our GCN-LSTM model against three baseline models: the conventional LSTM without dimension reduction, ConvLSTM, and CAE-LSTM . LSTM and ConvLSTM have been previously applied in the context of wildfire prediction and meteorological data forecasting~\cite{shi2015, liang2019}. For a deeper dive into the workings of CAE-LSTM, readers can refer to the related work section.

For training and validation, we utilise four out of the five ensemble members of the simulation, namely \(P_1\), \(P_2\), \(P_3\), and \(P_5\), while the remaining ensemble member \( P_4\) is used for testing. The testing approach is iterative. Starting with 12 months of unseen climate and wildfire data from the year 1961, we predict each subsequent 12 months of wildfire data until 1990. CAE-LSTM and ConvLSTM also follow this yearly sequence to sequence predictive method. Since our model, like the INFERNO model, focuses on predicting the fire burnt area \( P \), the remaining four input weather features are sourced from historical real weather data. These are concatenated with our \( P \) for each subsequent prediction. This approach is consistently applied across all models to ensure fairness.

Our model is implemented using PyTorch~\cite{paszke2019} and trained on an NVIDIA Tesla T4 GPU. The optimisation process employs the Adam optimiser~\cite{kingma2015} with a learning rate of 0.001. The objective was to minimise the mean squared error (MSE). The model is trained for 1000 epochs with a batch size of 12. We experimented with varying hyperparameters: the number of hidden layers in the range [1, 2] and the number of neurons in each hidden layer among [16, 32, 64]. To prevent overfitting, early stopping was incorporated. Hyperparameters were tuned based on the validation dataset performance, selecting the model configuration that yielded the lowest error.

The performance of each model on the test dataset was evaluated using several metrics, including MSE, Relative Root Mean Squared Error (RRMSE), Structural Similarity Index Measure (SSIM), and Peak signal-to-noise ratio (PSNR). Given our prediction target \( \mathbf{P}_T \in \mathbb{R}^{N \times 1 \times T} \), these metrics are defined as:

\begin{align}
\text{MSE} &= \frac{1}{N \times T} \sum_{i=1}^{N} \sum_{t=1}^{T} (\mathbf{P}_{T_{i,t}} - \hat{\mathbf{P}}_{T_{i,t}})^2 \\
\text{RRMSE} &= \frac{\sqrt{\text{MSE}}}{\bar{\mathbf{P}_T}} \\
\text{SSIM} &= \frac{(2\mu_{\mathbf{P}_T}\mu_{\hat{\mathbf{P}_T}} + c_1)(2s_{\mathbf{P}_T\hat{\mathbf{P}_T}} + c_2)}{(\mu_{\mathbf{P}_T}^2 + \mu_{\hat{\mathbf{P}_T}}^2 + c_1)(s_{\mathbf{P}_T}^2 + s_{\hat{\mathbf{P}_T}}^2 + c_2)} \\
\text{PSNR} &= 10 \cdot \log_{10}\left(\frac{\text{MAX}_{\mathbf{P}_T}^2}{\text{MSE}}\right)
\end{align}

Where \( \hat{\mathbf{P}_T} \) represents the predicted values, \( \mu \) and \( s \) denote the mean and standard deviation, respectively, and \( c_1 \) and \( c_2 \) are constants to stabilise the division with a weak denominator. To facilitate metric computations and image-based result presentations, we expanded \( \mathbf{P}_T \) through a decompression process using a land mask. This transformation yields a global wildfire burnt area image snapshots in dimensions \( T \times 112 \times 192 \). In this transformation, the non-land regions of the land-only column matrix \( \hat{\mathbf{P}_T} \) are filled with zeros, converting it into a complete image format. This specific step of ‘inflating’ the data with zeros is uniquely applied post-prediction for the GCN-LSTM model. On the other hand, the CAE-LSTM and Conv-LSTM models necessitate the inclusion of these zero-filled non-land regions from the start, meaning that their data needs to be in image format before the training process begins. Such practice of pre-filling with arbitrary values for model training might introduce bias, representing a potential drawback of these models.

\section{Numerical Results}\label{sec:results}

\FloatBarrier

\begin{table}[htb]
    \centering
    \caption{Performance metrics for models on predicting the Burnt area from 1961-1990.}
    \begin{tabular}{lcccc}
    \toprule
    Metrics & LSTM & Conv-LSTM & CAE-LSTM & GCN-LSTM \\
    \midrule
    MSE     & 0.001505 & 0.000723 & 0.000988 & \textbf{0.000498} \\
    RRMSE   & 0.506152 & 0.349927 & 0.409150 & \textbf{0.288990} \\
    SSIM    & 0.926949 & 0.950138 & 0.949862 & \textbf{0.966979} \\
    PSNR    & 26.784032 & 31.541234 & 29.348347 & \textbf{33.366938} \\
    \bottomrule
    \end{tabular}
    \label{tab:performanceMetrics}
\end{table}

\begin{figure}[htb]
    \centering
    \includegraphics[width=0.8\linewidth]{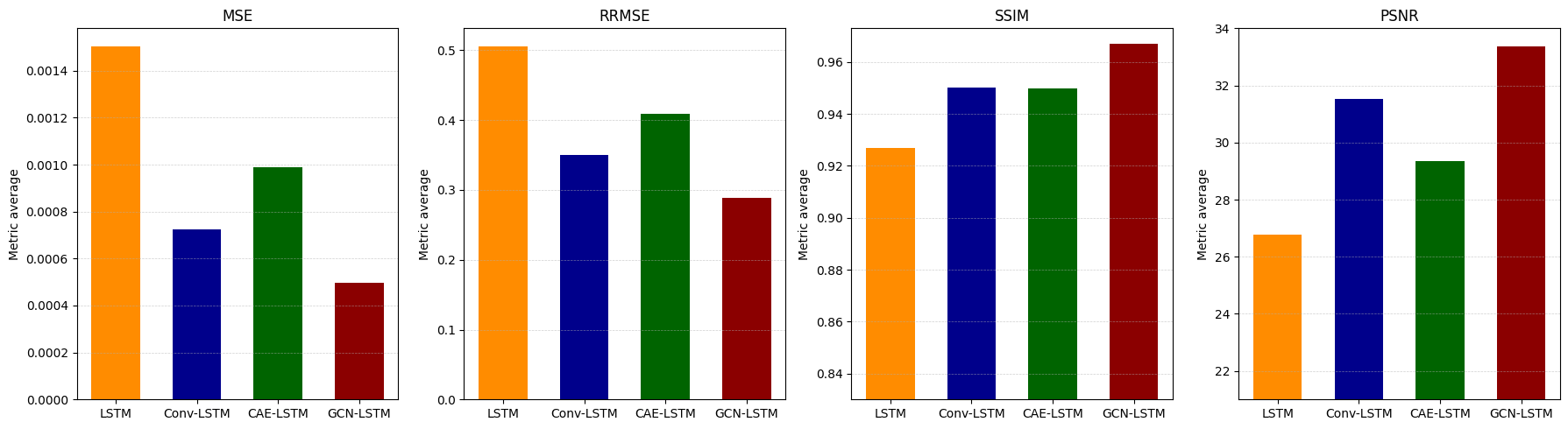}
    \caption{Bar chart of the overall performances of the models from 1961-1990.}
    \label{fig:barChartComparison}
\end{figure}

\begin{figure}[htb]
    \centering
    \includegraphics[width=0.8\linewidth]{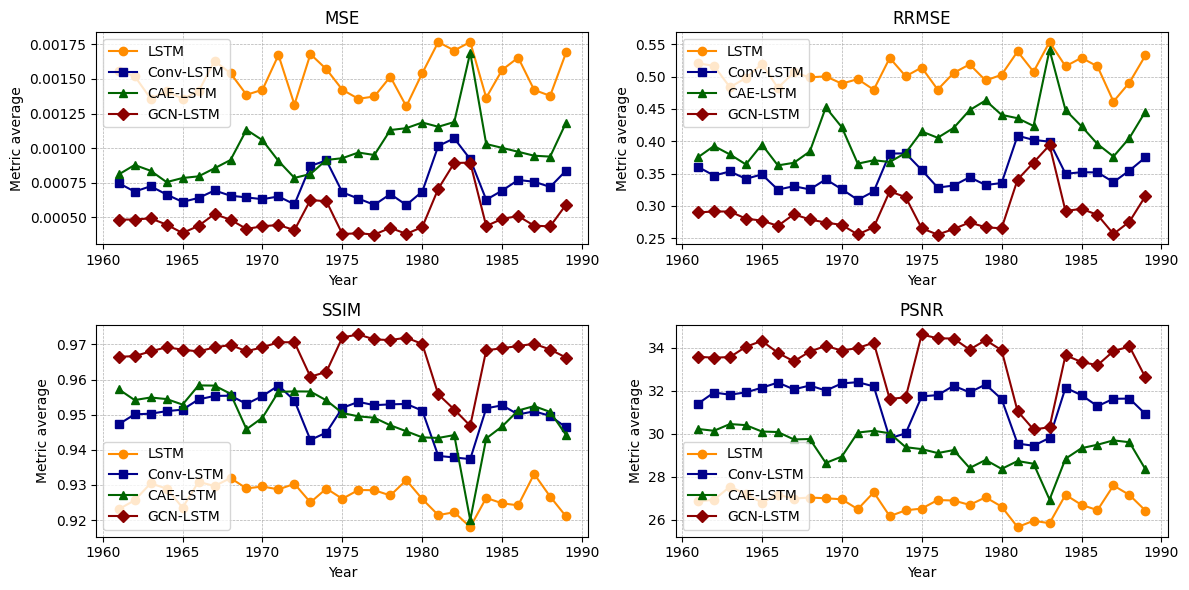}
    \caption{Yearly comparison of model performances from 1961-1990.}
    \label{fig:yearlyComparison}
\end{figure}
  
\begin{figure}[htbp]
    \centering
    \begin{subfigure}[b]{0.9\linewidth}
        \includegraphics[width=\linewidth]{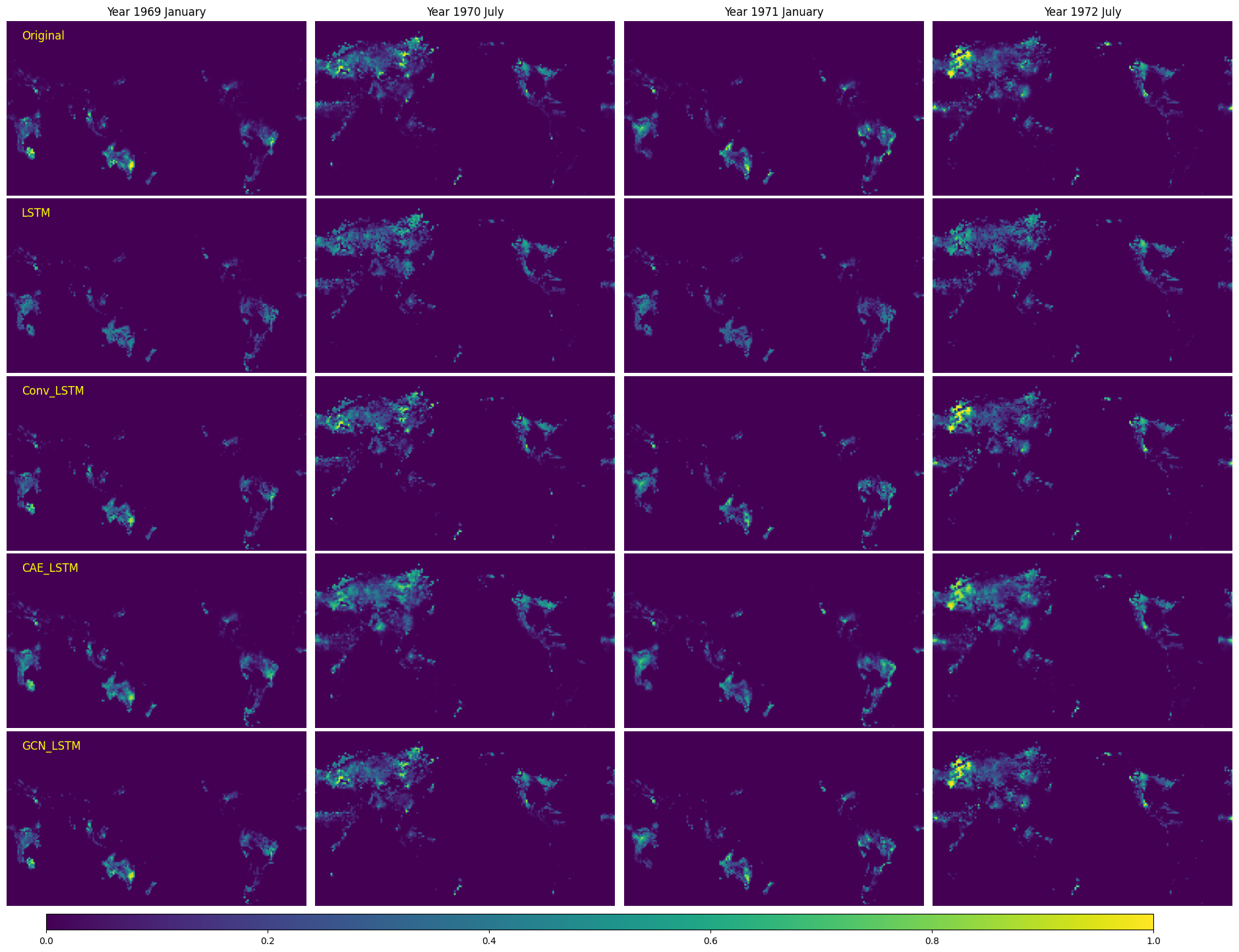}
        \caption{Year 1969-1972}
    \end{subfigure}
    \hfill
    \begin{subfigure}[b]{0.9\linewidth}
        \includegraphics[width=\linewidth]{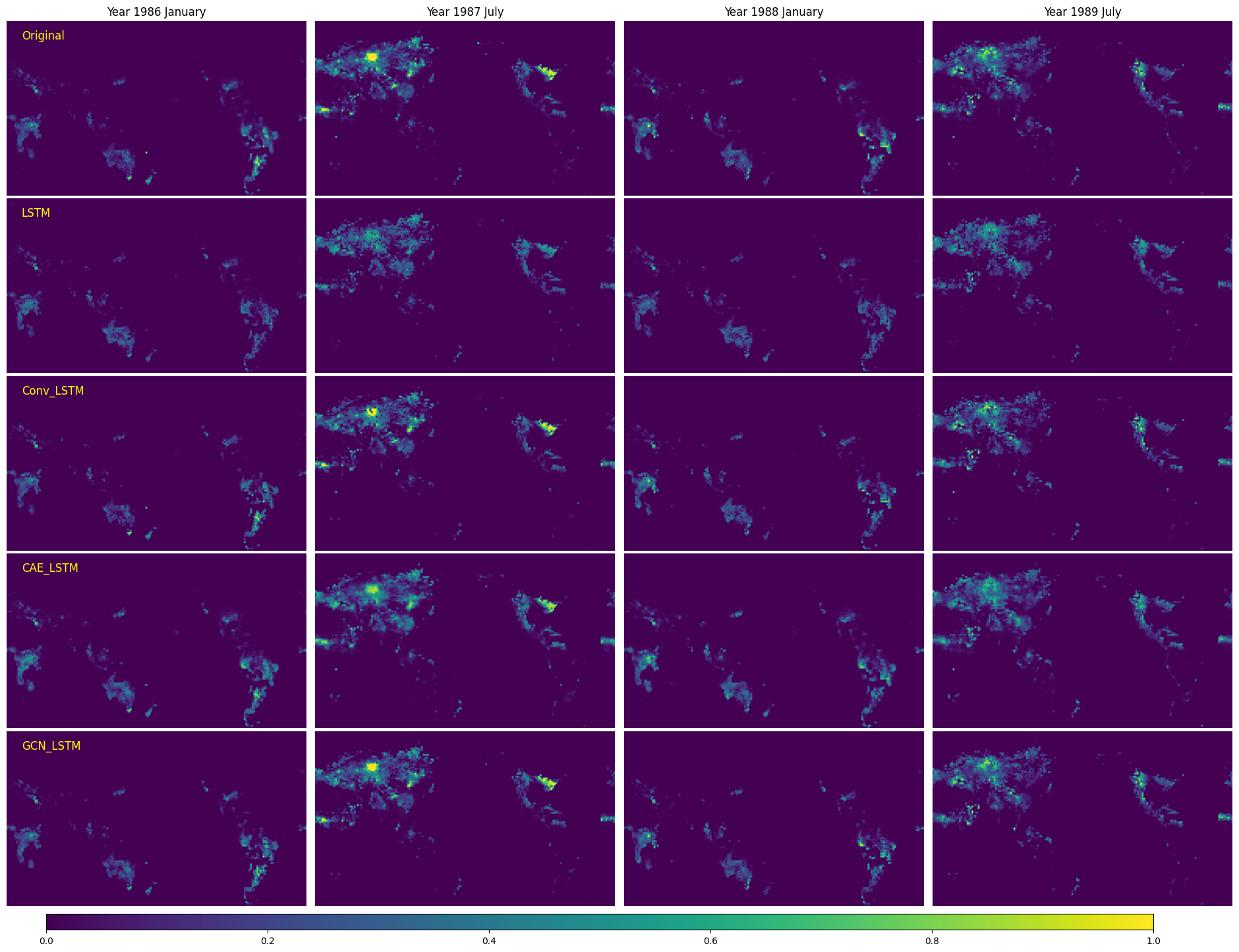}
        \caption{Year 1986-1989}
    \end{subfigure}
    \caption{Normalised original and predicted Burnt area fractions}
\label{fig:normalisedBurnt}
\end{figure}

\begin{figure}[htbp]
    \centering
    \begin{subfigure}[b]{0.9\linewidth}
        \includegraphics[width=\linewidth]{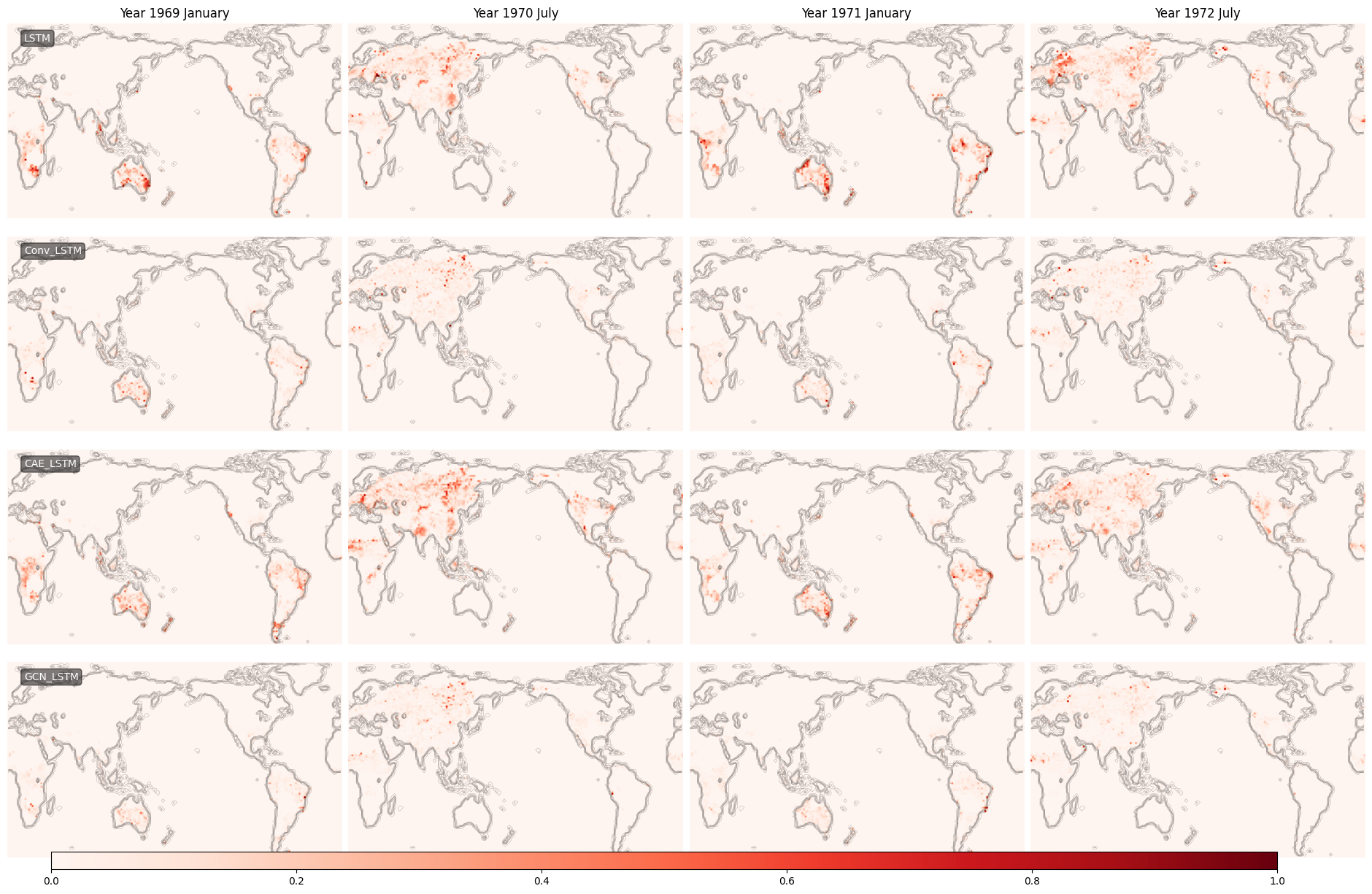}
        \caption{Year 1969-1972}
    \end{subfigure}
    \hfill
    \begin{subfigure}[b]{0.9\linewidth}
        \includegraphics[width=\linewidth]{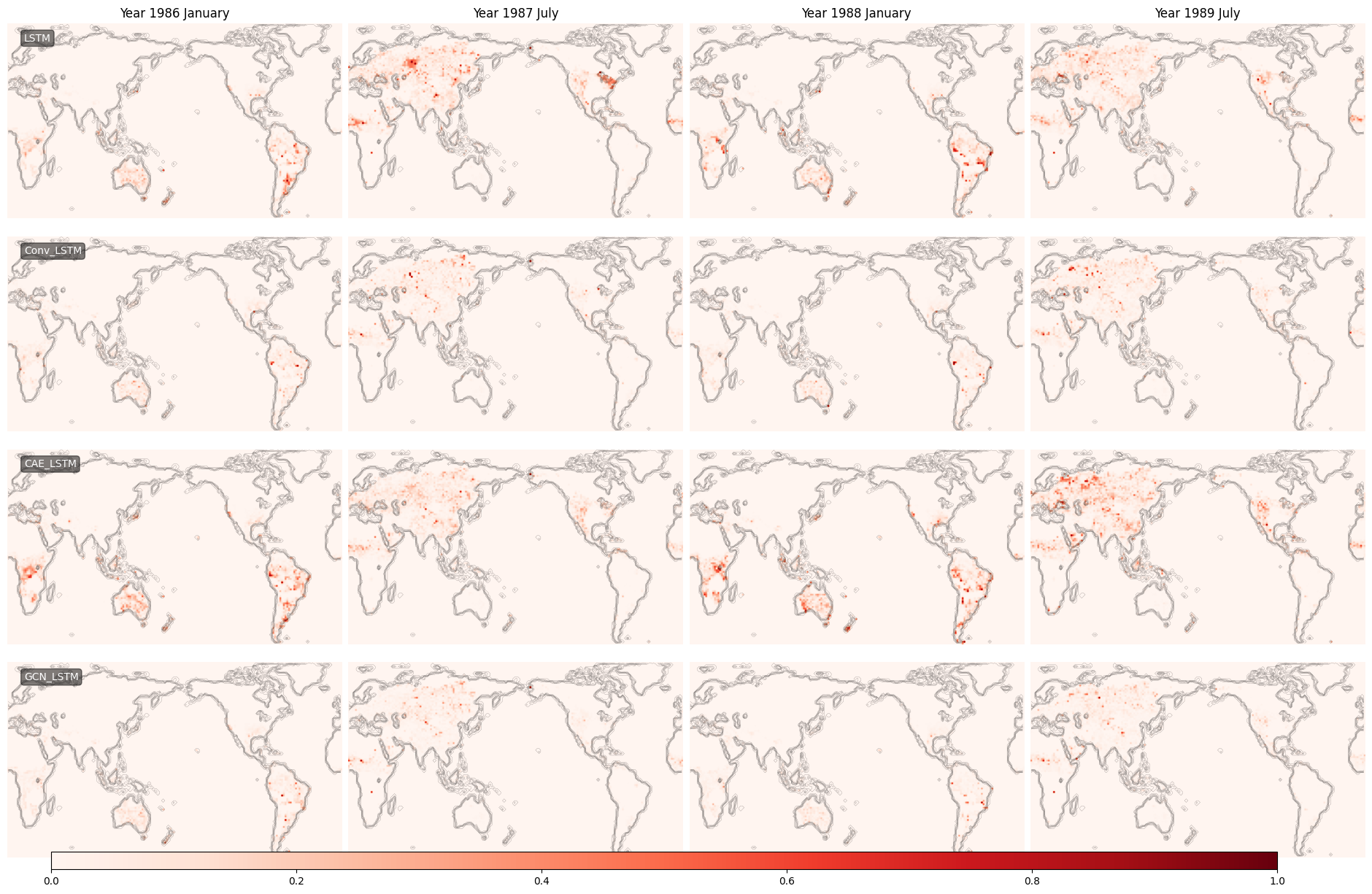}
        \caption{Year 1986-1989}
    \end{subfigure}
    \caption{Normalised absolute error of predicted against original Burnt area values}
\label{fig:differenceReal}
\end{figure}

Table~\ref{tab:performanceMetrics} summarised the performance of the GCN-LSTM model and three baseline models. For a visual representation of the above metrics in Table~\ref{tab:performanceMetrics}, we referred to the bar chart in Figure~\ref{fig:barChartComparison}. The GCN-LSTM model exhibited the best performance among the compared methods, consistently outperforming the baseline models across all metrics. More specifically, the model demonstrated the lowest MSE and RRMSE values, suggesting improved accuracy and error rates, respectively. Additionally, with the highest SSIM and PSNR values, GCN-LSTM provided superior structural similarity and peak signal-to-noise ratio performances.

Over a period of 30 years, as illustrated in Figure~\ref{fig:yearlyComparison}, both Conv-LSTM and CAE-LSTM demonstrated superior performance compared to the conventional LSTM model. Conv-LSTM generally outperformed CAE-LSTM throughout most of this timeframe; however, the performance gap between them was not substantial, with CAE-LSTM occasionally surpassing Conv-LSTM in certain years.  The GCN-LSTM model, however, consistently outperformed all the other models. As shown in Figure~\ref{fig:normalisedBurnt}, all models successfully predicted that wildfires predominantly occurred in the northern hemisphere during July and in the southern hemisphere during January, reflecting the seasonal nature of these wildfires~\cite{Pan2023WildfireSeason}. All models accurately captured the general regions affected by wildfires, with each predicting a certain extent of wildfire activity in the expected areas. Nevertheless, the GCN-LSTM model provided a more precise prediction of the wildfire's intensity in these regions. Figure~\ref{fig:differenceReal} illustrates the visual comparisons of the absolute errors in all the models' predictions against the ground truth. Both Conv-LSTM and CAE-LSTM exhibit better alignment with the ground truth than the conventional LSTM model, with Conv-LSTM delivering a superior performance compared to CAE-LSTM. The superior spatial precision of the GCN-LSTM is confirmed, showcasing minimal discrepancies between its predictions and the actual data among all models. This evidence collectively highlights the robustness and effectiveness of the GCN-LSTM model, solidifying its status as a leading model in terms of predictive accuracy and reliability.

\FloatBarrier

\section{Explainability}\label{sec:explainability}

This section explores the key aspects that bolster our model's interpretability: the inherent characteristics of the GNN, insights from community detection, and the significance of feature and node attributions.

\subsection{Community Detection}
\label{sec:community}
Our model's enhanced interpretability is rooted in its graph representation, which addresses the null data locations issue prevalent in image snapshot representation used in CNN-based models. Additionally, explicit modelling of information flow through graph edges~\cite{kipf2017semisupervised} provided more interpretable prediction results. The graph edges, constructed based on wildfire occurrence correlation \( r \), offer a novel representation of global climate and wildfire data.

To distinguish potential clusters or regions with wildfire occurrence correlations, we applied community detection on the constructed graph. Community detection aims to identify densely connected groups of nodes in a graph, with the nodes within a group being more connected to each other than nodes outside the group. Specifically, the Louvain method was employed for this purpose~\cite{louvain}. The Louvain method is a widely-used heuristic for modularity optimisation. The Louvain method optimises the modularity \( Q \) of a partition, defined as:

\begin{equation}
Q(\gamma) = \frac{1}{2m} \sum_{ij} \left( A_{ij} - \gamma \frac{k_i k_j}{2m} \right) \delta(c_i, c_j) ,
\end{equation}
\begin{equation}
k_i = \sum_{j} A_{ij},
\end{equation}


where \( m \) is the number of links, \(A_{ij}\) are the elements of the adjacency matrix obtained using wildfire occurrence correlation \( r \), as defined in Equation~\ref{eq:adjancency}, \( k_i \) is the sum of the weights of the links attached to node \( i \), \( c_i \) is the community to which node \( i \) is assigned, and \( \delta(u, v) \) is 1 if \( u = v \) and 0 otherwise. \( \gamma \) is the resolution parameter, which determines the granularity of detected communities. A larger value of \( \gamma \) leads to smaller communities, while a smaller value results in larger communities.

\begin{figure}[htb]
\centering
\includegraphics[width=0.69\linewidth]{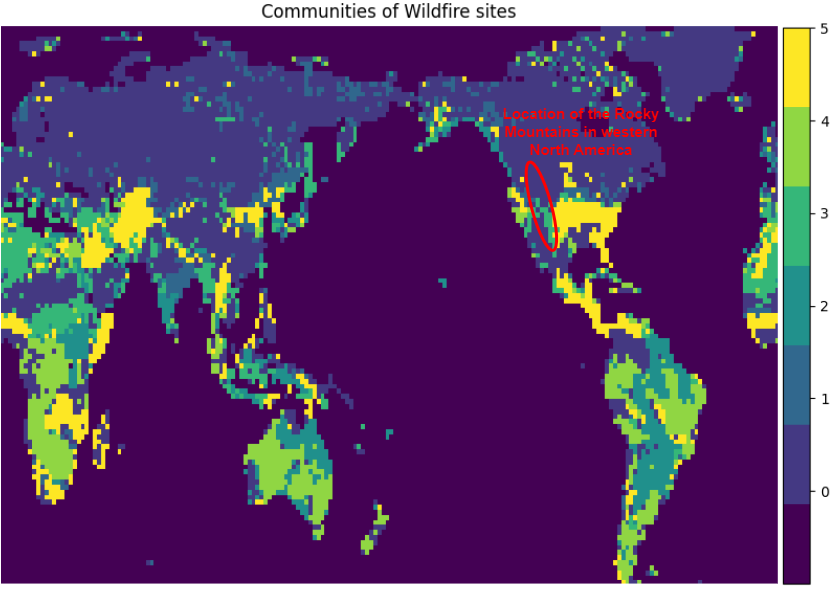}
\caption{Community detection results using the Louvain method at \( \gamma \) of 1.06. Different colors represent different communities}
\label{fig:communities}
\end{figure}

Our community detection results obtained using the Louvain method with a resolution parameter at 1.06, revealed intriguing patterns, as depicted in Figure~\ref{fig:communities}. Communities were predominantly located in specific hemispheres; each one either concentrated mainly north or mainly south of the equator. This distribution pattern aligns with the wildfire seasonality, given that wildfires primarily occur during the summer months in each hemisphere~\cite{Pan2023WildfireSeason}. Additionally, vast regions including parts of the Sahara Desert, parts of the Arabian Peninsula, and Greenland were absent from all communities apart from the default community, likely owing to the fact that the vegetation profiles of these areas might not support the occurrence of wildfires; this result is in line with the observed distribution of global wildfire fire under current conditions~\cite{Krawchuk2009}. The influence of natural barriers was also observed: regions separated by the Rocky Mountains in the USA were identified as distinct communities. Our results from community 5 (yellow contour in Figure~\ref{fig:communities}), particularly the geographic areas concerning the eastern USA and Central America, intriguingly largely align with the findings of Archibald et. al~\cite{pyromes2013}. In their study, a Bayesian clustering algorithm was used to discern five global fire regime pyromes. Even with their approach grounded on more comprehensive tabular data and with several additional characteristics to consider, the regions they categorised under the intermediate–cool–small (ICS) pyrosome bear similarities to our community detection results of community 5. The consistency of our model's findings with previous research emphasises its interpretability. This consistency indicates that the model is able to comprehend the underlying relationships and dynamics in the data and is not just overfitting a specific pattern. 

We employ community detection analysis to identify potential clusters associated with wildfire occurrences. These identified clusters closely align with both our expectations and findings from prior research. Interestingly, some clusters span across vast oceans, indicating that our model is able to process both proximity and long-range information, and addresses the challenges of long-range correlations. Such results underline the rationale behind our graph-based computations. To further substantiate this assertion, we will validate this with global node attribution on our graph.

It's critical to note that our ground truth data, derived from JULES-INFERNO under LGM climatic conditions, were considerably colder than today's climate. The LGM-based data might exhibit discrepancies from historical data, especially in wildfire locations. Thus, while our insights offer an academic perspective, it's important to approach their real-world applicability judiciously. With contemporary data inputs, however, our model's relevance could be further realised.


\subsection{Feature Importance}

To elucidate the model's decision-making process, we employed the Integrated Gradient (IG) method~\cite{integratedgradient}. The IG method had been widely adopted to interpret neural networks in geographical-related forecasting tasks, including feature importance and sensitivity analysis~\cite{Lu2021NeuralNI, Kratzert2021}. This method provides a measure of feature importance by attributing the prediction of a deep learning model to its input features. The attribution of the \( i_{th}\) feature, \( \mathcal{G}_i\), is computed as the path integral of the straight-line path from the baseline input matrix, \( \bar{X} \), to the input, \( X \), formalised as:

\begin{equation}
\mathcal{G}_i = (X_i - \bar{X}_i) \times \int_0^1 \frac{\partial F(\bar{X} + \alpha \times (X - \bar{X}))}{\partial X_i} d\alpha,
\end{equation}

where \( X \) represents the input feature matrix to the model, encompassing both the original fire and climate data feature inputs, as defined in Equation~\ref{eq:feature_matrix}. \( \bar{X} \) serves as the baseline input matrix, populated entirely with zero data points, providing a reference point for comparison and helping to isolate and understand the specific impact of each feature from the input matrix 
\( X \) on the model’s predictions. The scalar \( \alpha \), which ranges between 0 and 1, facilitates the interpolation between \( \bar{X} \) and \( X \). When \( \alpha = 0 \), the matrix aligns with the baseline \( \bar{X} \), and at \( \alpha = 1 \), it coincides with the actual input matrix \( X \). By varying \( \alpha \), matrices can be produced that manifest as combinations of both \( \bar{X} \) and \( X \). This interpolation method enables a systematic examination of the model's predictions as one incrementally integrates features from the baseline to the actual input. Both \( \alpha \) and \( \bar{X} \) play key role in understanding how removing or changing each pixel affects the predictions made by the model. In essence, IG method serves as a tool to examine the influence of each input feature on a deep learning model's output, highlighting the importance of specific features in the model’s decision-making process.

\begin{figure}[ht]
\centering
\begin{subfigure}{.5\textwidth}
  \centering
  \includegraphics[width=1.0\linewidth]{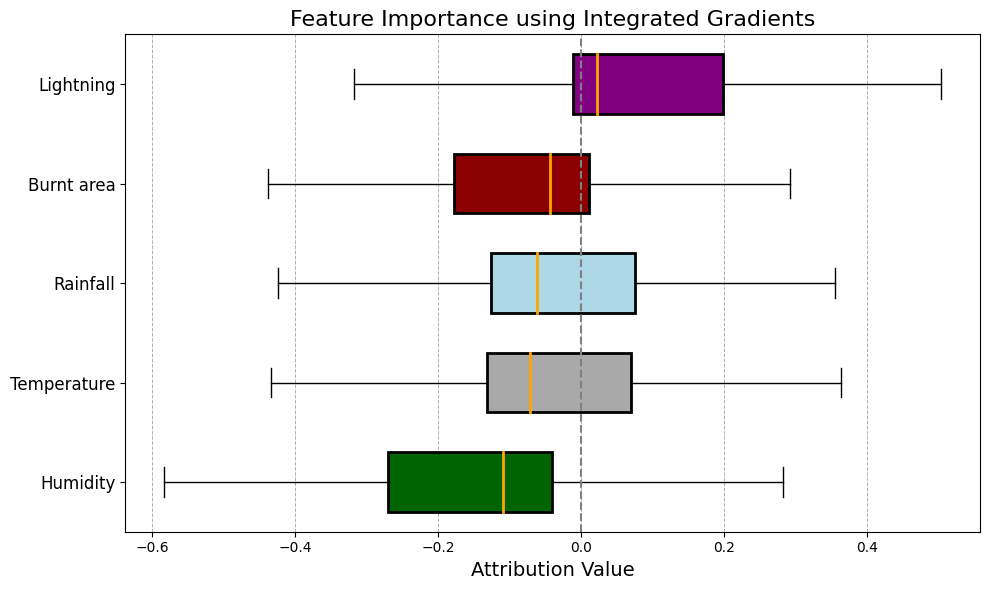}
  \caption{Attributions for prediction 1 month into future}
  \label{fig:boxplot2}
\end{subfigure}%
\begin{subfigure}{.5\textwidth}
  \centering
  \includegraphics[width=1.0\linewidth]{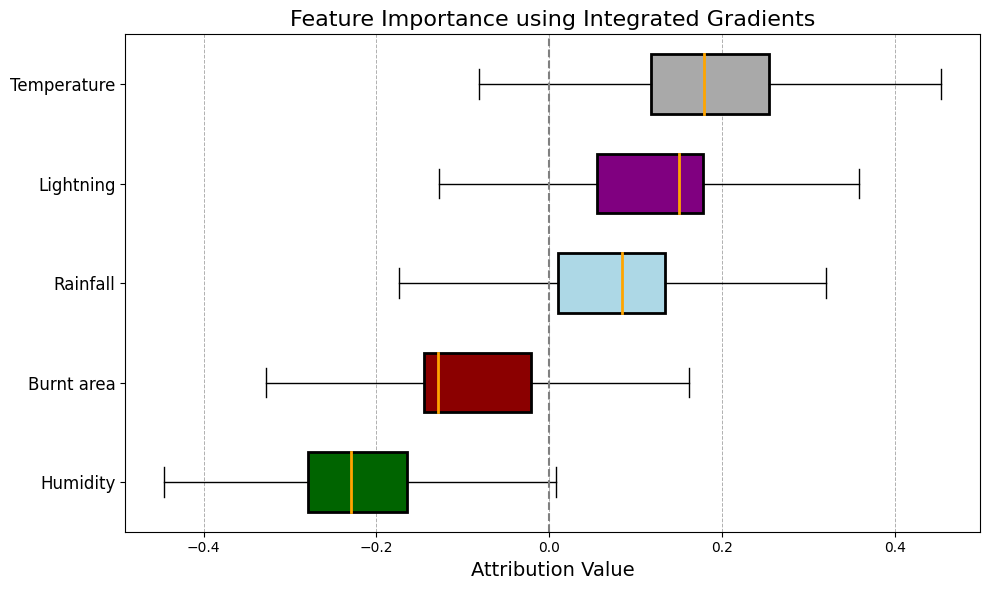}
  \caption{Attributions for prediction 11 month into future}
  \label{fig:boxplot1}
\end{subfigure}
\caption{Feature attributions using Integrated Gradients for different prediction horizons.}
\label{fig:boxplots}
\end{figure}

For our analysis, we sampled 500 random nodes from our graph and used their aggregated data to generate the box plots in Figure~\ref{fig:boxplots}. These plots provide insights into the importance of each feature for predictions at different forecast horizons. As illustrated in (Figure~\ref{fig:boxplot2}) and (Figure~\ref{fig:boxplot1}), both humidity (\( Hum_t \)) stands out with a pronounced negative attribution, aligning with the understanding that higher humidity levels are unfavourable to wildfire spread. To put our findings in context, we draw comparisons with the work of Kondylatos et al., who carried out a SHAP analysis on a 15-variable-driven deep learning wildfire prediction model and found that soil moisture and relative humidity were the two most significant negative contributors to fire danger~\cite{kondylatos2022wildfire}, it becomes evident that our analysis is consistent with these earlier observations, emphasising the reliability and validity of our approach.

The model's attention to features also varies depending on the prediction horizon. For the prediction of wildfire occurrences in the upcoming month (Figure~\ref{fig:boxplot2}), lightning (\( L_t \)) emerges as the feature with the highest positive attribution. In other words, the model pays more attention to lightning when it comes to predicting wildfires that are expected to happen in the next month. In contrast, when the prediction extends to a more distant future, specifically 11 months ahead (Figure~\ref{fig:boxplot1}), temperature (\( T_t \)) takes precedence with the highest positive attribution, followed by lightning (\( L_t \)). Extremely hot temperatures in dry conditions and lightning are known to be closely associated with increasing wildfire danger, particularly as this prediction horizon almost completes a full seasonal cycle, implying that the climatic conditions influencing wildfires at this future point are expected to be similar to the present, thus aligning our findings well with existing wildfire research~\cite{kondylatos2022wildfire, ORDONEZ201244, perez-invernon2023variation}. Additionally, This shift in feature importance underscores the model's adaptability in weighing features differently based on the temporal context.

In essence, these insights into feature importance not only enhance the transparency of our model but also provide a deeper understanding of the factors driving wildfire predictions for different forecast horizons. 

\subsection{Node Importance}

In addition to recognising feature significance, understanding the influence of individual nodes on model predictions plays a crucial part in explaining our model. This nodewise analysis offers a spatial perspective on how different regions in our graph contribute to specific predictions. For this, we leveraged the IG method to compute the attribution across all nodes for selected node predictions, thereby performing a thorough and detailed nodewise analysis.

To illustrate, we selected two nodes based on the communities identified in the earlier community detection subsection. These nodes belong to communities 5 and 4, respectively, each predominantly concentrated in a different hemisphere, thus providing a broad geographic representation. The resulting attention maps, which highlight important regions in data that a model focuses on when making a decision, visualise the significance of various nodes in making predictions for the selected nodes.

\begin{figure}[ht]
    \centering
    \begin{subfigure}[b]{0.49\textwidth}
        \includegraphics[width=\textwidth]{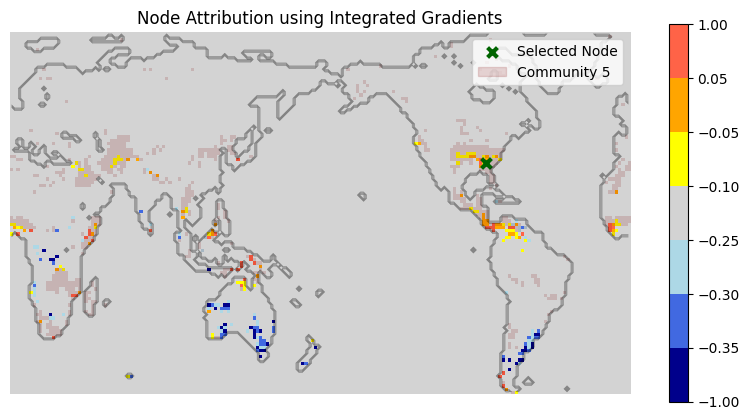}
        \caption{Node from Community 5}
        \label{fig:node5}
    \end{subfigure}
    \hfill
    \begin{subfigure}[b]{0.49\textwidth}
        \includegraphics[width=\textwidth]{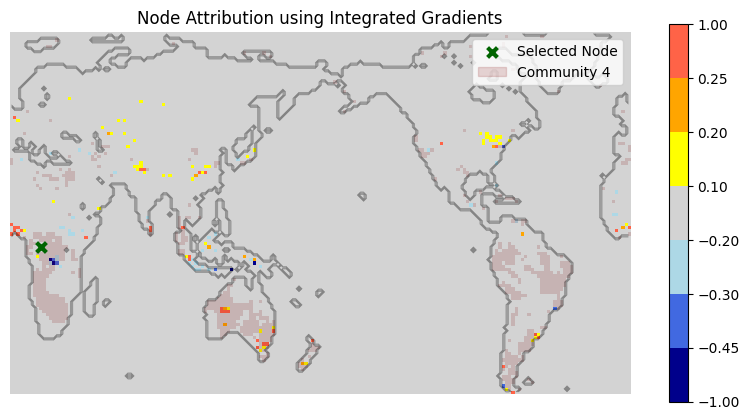}
        \caption{Node from Community 4}
        \label{fig:node4}
    \end{subfigure}
    \caption{Node attributions for selected nodes in communities 5 and 4 using the IG method.}
    \label{fig:node_attributions}
\end{figure}

From Figure~\ref{fig:node_attributions}, we observe a natural expectation: the most pronounced positive and negative contributions generally arise from geographically proximate regions. However, the strength of our graph representation shines through as even distant regions, across different continents display notable influence.

Specifically, examining Figure~\ref{fig:node5}, the node located in the eastern US within Community 5, reveals intriguing patterns. Apart from its immediate surroundings, regions in Central America and the African coast, despite being geographically distant, show significant positive contributions. These regions are also encompassed within Community 5, echoing our findings from Section~\ref{sec:community}. Notably, these observations closely align with the ICS wildfire pyrosome as described by Archibald et al.~\cite{pyromes2013}. These results confirm that the model's predictions extend beyond immediate proximities. Distant regions, when interconnected through climatic or wildfire patterns captured in our graph representation, play an integral role in the decision-making process. This indicates that our graph-based model can effectively utilise long-range information, overcoming the limitations of long-range dependencies inherent in traditional CNN-based methods.

\subsection{Summary \& Discussion}

\begin{itemize}

\item \textbf{Community Detection}: 
Our community detection results, obtained using the Louvain method, identify potential clusters associated with wildfire occurrences. Certain identified clusters largely align with prior research, signifying the robustness and relevance of our findings. Some clusters span multiple continents, indicating that our model can process both proximate and long-range information, addressing the challenges of long-range correlations, and reinforcing the logic of our graph-centric approach.

\item \textbf{Feature Importance}: 
Using the IG method, key features such as humidity, lightning, and temperature emerged as prominent in determining wildfire dynamics at different forecast horizons, with their varying importances indicating the model's adaptability to temporal contexts. The alignment with established major factors from prior research contributing to wildfire danger highlights the model's ability to extract meaningful insights from the data.

\item \textbf{Node Importance}: 
We present the nodewise analysis on how different regions contribute to specific predictions, notably the attention maps obtained using the IG method. The significant influence exerted by both nearby and distant nodes in our model's predictions emphasises its ability to process intricate data relationships. Notably the capability to attribute importance to geographically distant nodes that share climatic or wildfire patterns, illustrating the model's edge in leveraging long-range information over traditional CNN approaches. 

\end{itemize}

In summary, this section emphasises the strengths of our GNN-based model in terms of capturing long-range dependencies, its adaptability to temporal contexts, and its ability to deliver insights that are consistent with prior research, highlighting its ability to overcome the limitations faced by traditional CNN-based methods.

\section{Conclusion}\label{sec:conclusion}

To conclude, we introduced a novel GNN-based model for global wildfire prediction, leveraging the strengths of both Graph Convolutional Networks (GCNs) and Long Short-Term Memory (LSTM) networks. By transforming global climate and wildfire data into a graph representation, we effectively addressed challenges such as null oceanic data and long-range dependencies, which have been persistent issues in conventional CNN-based models.

Our model's performance, as benchmarked against established architectures using an unseen ensemble of JULES-INFERNO simulations from 1961-1990, demonstrated superior predictive accuracy. Notably, the GCN-LSTM model consistently outperformed baseline models across various metrics, underscoring its efficacy in the domain of wildfire prediction.

Furthermore, the explainability of our model was highlighted through community detection, feature importance, and node attribution analyses. The feature importance analysis, conducted using the Integrated Gradient method, elucidated the model's decision-making process, highlighting the significance of various meteorological factors at different prediction horizons. The community detection, based on the Louvain method, revealed potential clusters of regions with wildfire correlations. This understanding was deepened when specific nodes from these communities were examined through node attribution analysis, confirming the model's adeptness at assimilating information from both proximate and geographically distant, yet climatically correlated regions.

In conclusion, our research not only advances the methodological landscape of wildfire prediction but also emphasises the importance of model transparency and interpretability. The insights derived from our model can serve as valuable tools for stakeholders in wildfire management and prevention, aiding in the development of more informed and effective strategies. Moving forward, a promising direction for future research involves training and testing our model on real-world data, as this work is based on simulations under the LGM meteorological conditions. Employing data assimilation techniques to integrate observed data into the model could significantly enhance its accuracy and reliability, making it a more robust tool for wildfire prediction. Additionally, while our current model primarily focuses on predicting the wildfire burnt area variable, there is substantial potential to extend its capabilities to encompass full climate modelling. This expansion would not only provide a more comprehensive understanding of wildfire dynamics but also contribute to the broader field of remote sensing, meteorology and climate science, offering invaluable insights for climate change mitigation and adaptation strategies.

\section*{Data and code availability}
The code and data of this work is available at \url{https://github.com/DL-WG/gcn-lstm-wildfire}.

\section*{Acknowledgements}

 This research is funded by the Leverhulme Centre for Wildfires, Environment and Society through the Leverhulme Trust, grant number RC-2018-023. JULES-INFERNO simulations were carried out using JASMIN, the UK's collaborative data analysis environment (https://jasmin.ac.uk)
 
\bibliographystyle{IEEEtran}
\footnotesize
\bibliography{main}

\end{document}